# Illuminant Spectra-based Source Separation Using Flash Photography

Zhuo Hui,[†] Kalyan Sunkavalli,[‡] Sunil Hadap,[‡] and Aswin C. Sankaranarayanan[†]
[†]Carnegie Mellon University    [‡]Adobe Research

## Abstract

*Real-world lighting often consists of multiple illuminants with different spectra. Separating and manipulating these illuminants in post-process is a challenging problem that requires either significant manual input or calibrated scene geometry and lighting. In this work, we leverage a flash/no-flash image pair to analyze and edit scene illuminants based on their spectral differences. We derive a novel physics-based relationship between color variations in the observed flash/no-flash intensities and the spectra and surface shading corresponding to individual scene illuminants. Our technique uses this constraint to automatically separate an image into constituent images lit by each illuminant. This separation can be used to support applications like white balancing, lighting editing, and RGB photometric stereo, where we demonstrate results that outperform state-of-the-art techniques on a wide range of images.*

## 1. Introduction

Real-world lighting often consists of multiple illuminants with different spectra. For example, outdoor illumination – both sunlight and skylight – differ in color temperature from indoor illuminants like incandescent, fluorescent, and LED lights. These variations in illuminant spectra manifest as color variations in captured images that are often a nuisance for vision-based analysis and photography applications.

In this work, we address the problem of explicitly separating an image into multiple images, each of which is lit by only one of the illuminants in the scene (see Figure 1(b)). Source separation of this form can enable a number of image editing and scene analysis applications. For example, we can change the illumination in the image by editing each illuminant image, or use the multiple images for scene analysis tasks like photometric stereo.

However, source separation is a highly ill-posed inverse problem. It is especially hard from a single photograph; each pixel observation in the image combines the effect of the unknown mixture of illuminants and the unknown scene reflectance. Previous attempts at addressing these challenges either use calibrated acquisition systems [14, 13] or rely on extensive user input [7, 6, 8], making it difficult to apply them at large-scale.

In this paper, we take a step towards source separation by making use of flash photography – two photographs acquired with and without the use of the camera flash – a light-weight capture setup that is available on most consumer cameras. The key insight behind our technique is that flash photography provides an image under a single illuminant, thereby enabling us to infer the reflectance spectra up to a per-pixel scale. Based on this, we derive a novel reflectance-invariant — the *Hull Constraint* — that relates light source spectra and their relative per-pixel shading to the observed intensities in the no-flash photograph. We use the Hull Constraint to separate the no-flash photograph into multiple images – each corresponding to the lighting of a unique spectra. This, in turn, enables a wide-range of capabilities including white-balancing under complex mixed illumination, the editing of the color and brightness of the separated illuminants, camera spectrum response editing and photometric stereo. The Hull constraint is independent of scene and lighting geometry; it applies equally to point and area sources as well as near and distant lighting. Figure 1 showcases our technique for a real-world sample.

**Contributions.**   We propose a flash photography-based algorithm to analyze spatially-varying, mixed illumination. In particular, we make the following contributions:

1. We introduce a novel reflectance-invariant property of Lambertian scenes that relates illuminant spectra to observed pixel intensities.

2. We propose an algorithm to separate an image into its constituent illuminants, and present an analysis of the robustness and limitations of this technique.

3. We leverage these separated images to enable a wide variety of applications including white balancing, light editing, camera response editing and photometric stereo.

**Assumptions.**   The techniques proposed in this paper assume that the imaged scene is Lambertian and that scene reflectance and illuminant spectra lie in a low-dimensional subspace. We assume that the flash illuminates all the scene points and that the flash/no-flash images are aligned. The success of our technique to resolve source separation also relies on the existence of pure pixels as defined later.

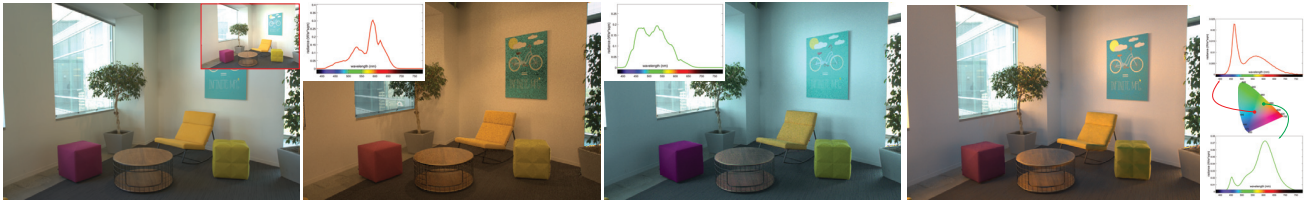

(a) No-flash / flash images      (b) Source separation results      (c) Illumination editing results

Figure 1. The scene in (a) is lit by cool sky illumination from the window on the left and warm indoor lighting from the top. Given a pair of no-flash/flash images, our method separates the no-flash image into two images lit by each of these illuminants (b) and estimates their spectral distribution (insets in (b)). Using our illuminant estimates, we are able to edit the illumination in the photograph (c) by changing the individual spectra of the light sources (insets in (c)).

## 2. Related Work

In this section, we review previous works on illumination analysis as well as prior applications of flash photography.

### 2.1. Lighting analysis

**Active illumination.** Active illumination methods use controlled illumination to probe and infer scene properties. Controlled capture setups like a light stage [14] capture images of a person or a scene under all lighting directions and re-render photorealistic images under arbitrary illumination [16, 36]. Another class of techniques rely on projector-camera systems to probe and separate light transport in a scene [33]. While active techniques can enable high-quality illumination analysis and editing, these systems are complex and expensive. In contrast, we propose a simple capture process that uses a camera flash to enable a number of illumination analysis, editing and reconstruction tasks.

**Passive illumination.** Passive illumination methods aim to estimate scene properties from images captured as-is under natural illumination. Barron and Malik [4, 5] demonstrate impressive results for shape, reflectance, and illumination estimation for a single object captured under low-frequency distant lighting. Johnson and Malik [30] use spectral variations in real-world illumination to reconstruct shape from shading. These methods rely on scene priors that are often violated on real-world scenes with complex geometry, reflectance, and spatially-varying lighting. In contrast, we demonstrate that the use of flash photography can lead to high-quality lighting (and shape) estimates without the same restrictive assumptions. Recently, deep learning-based methods have been proposed to infer illumination [26, 22] from a single image. However, these methods do not support pixel-level image editing, which our method does by explicitly separating an input image into its constituent components.

**Color constancy.** Color constancy — the problem of correcting for the spectrum of the scene illumination — is a closely related light analysis problem, and has been extensively studied [23]. Previous work models the effect of changing the illumination spectral distribution as a (typically linear) transformation of the observed pixel intensities. The seminal work of Finlayson et al. [20, 19] demonstrates real-world reflectance and illumination spectra lie in low-dimensional spaces, allowing for the use of a diagonal transformation. Chong et al. [12] build on this to derive conditions for the basis which can "best" support diagonal color constancy. Current color constancy methods range from physics/low-level feature-based methods [43, 15, 24] to learning-based approaches [31, 10] to user-driven interactive solutions [27, 8]. The vast majority of these methods assume a single illuminant in the scene. While our approach is built on top of diagonal color constancy techniques, we can handle multiple illuminants and can go beyond color constancy and separate the captured image into constituent images lit by individual illuminants.

### 2.2. Flash photography

Flash photography refers to techniques that capture two images of a scene — with and without flash illumination. It has been used for image denoising [37, 17], deblurring [45], artifact removal [17], non-photorealistic rendering [39], foreground segmentation [41] and matting [42]. More recently Hui et al. [28] propose a flash photography-based white balancing method for mixed illumination. Our approach is similar in spirit to this work in that we leverage the flash/no-flash image pair to obtain a reflectance-invariant image that is used for lighting analysis. Going beyond that, we derive a novel physics-based constraint that enables explicit separation of the contribution of different light sources at each pixel. Our analysis enables a number of applications that are not possible with prior work [28], including light editing, and two-shot photometric stereo.

## 3. The Hull Constraint

Given an image of a scene lit by a mixture of illuminants — the *no-flash* image — our goal is to estimate the contribution of each illuminant to the observed pixel intensities. In this section, we set up the image formation model and derive a novel constraint between the observed no-flash/flash pixel intensities and the contributions of each scene illuminant to the scene appearance.

### 3.1. Problem setup and image formation

We assume that the scene is Lambertian and is imaged by a three-channel color camera. The intensity of the no-flash image observed at a pixel $\mathbf{p}$ in the $k$-the color channel ($k \in \{r, g, b\}$) is

$$I_{\text{nf}}^k(\mathbf{p}) = \int_\lambda \rho_{\mathbf{p}}(\lambda) S^k(\lambda) \ell_{\mathbf{p}}(\lambda) d\lambda, \quad (1)$$

where $\rho_{\mathbf{p}}$ is the reflectance spectra, $S^k$ is the camera spectral response for the $k$-th channel and $\ell_{\mathbf{p}}(\lambda)$ is the light spectra at pixel $\mathbf{p}$. When the scene is lit by $N$ light sources, the light spectra at pixel $\mathbf{p}$ can be expressed as

$$\ell_{\mathbf{p}}(\lambda) = \sum_{i=1}^{N} \eta_i(\mathbf{p}) \ell_i(\lambda),$$

where $\ell_i(\lambda)$ is the spectra of the $i$-th light source and $\eta_i(\mathbf{p})$ is the shading induced by the $i$-th source at pixel $\mathbf{p}$. The shading term $\eta_i(\mathbf{p})$ is assumed to be non-negative. Note that $\eta_i(\mathbf{p})$ does not have a simple analytical form, thereby can accommodate point, extended and area light sources. The illumination spectra $\{\ell_1, \ldots, \ell_N\}$ are assumed to be the same for all pixels and hence, any spatial light fall-off is captured in the shading term. With this, (1) can be written as

$$I_{\text{nf}}^k(\mathbf{p}) = \int_\lambda \rho_{\mathbf{p}}(\lambda) S^k(\lambda) \left( \sum_{i=1}^{N} \eta_i(\mathbf{p}) \ell_i(\lambda) \right) d\lambda. \quad (2)$$

Estimating the reflectance, shading and illumination parameters as well as separating the no-flash photograph into $N$ photographs — one for each of the $N$ light sources — are hard inverse problems. The parameters of interest, namely $\rho$ and $\ell_i$ are infinite-dimensional. Further, the multi-linear encoding of the reflectance, shading and illumination parameters in the image intensities leads to a highly-ambiguous solution space. To resolve these challenges, we make two key assumptions.

**Assumption 1 — Reflectance and illumination subspaces.** We assume that the reflectance and illumination spectra in the scene are well-approximated by low-dimensional subspaces. Given a reflectance basis $B_R = [\widetilde{\rho}_1(\lambda) \ldots \widetilde{\rho}_{M_1}(\lambda)]$ and an illumination basis $B_L = [\widetilde{\ell}_1(\lambda) \ldots \widetilde{\ell}_{M_2}(\lambda)]$, we can write

$$\rho_{\mathbf{p}}(\lambda) = B_R \, \mathbf{a}_{\mathbf{p}}, \quad \ell_i(\lambda) = B_L \, \mathbf{b}_i.$$

Here, $\mathbf{a}_{\mathbf{p}} \in \mathbb{R}^{M_1}$ are the reflectance coefficients at pixel $\mathbf{p}$ and $\mathbf{b}_i \in \mathbb{R}^{M_2}$ are the illumination coefficients for the $i$-th source. To resolve the ambiguity in the definition of the shading, we assume that the lighting coefficients are unit-norm, i.e., $\|\mathbf{b}_i\|_2 = 1$; hence, the illumination coefficients are points on the 2D sphere. Given this, we write (2) as:

$$I_{\text{nf}}^k(\mathbf{p}) = \sum_{i=1}^{N} \mathbf{a}_{\mathbf{p}}^\top E^k \eta_i(\mathbf{p}) \mathbf{b}_i. \quad (3)$$

Here, $E^k$ is the $M_1 \times M_2$ matrix defined as

$$E^k(i, j) = \int_\lambda \widetilde{\rho}_i(\lambda) S^k(\lambda) \widetilde{\ell}_j(\lambda) d\lambda.$$

Finally, as a consequence of having 3-color images, we will need to restrict $M_1 = M_2 = 3$. Real world reflectance and illumination spectra are known to be well-approximated by low-dimensional subspace — an insight that is used extensively in the color constancy [12, 20, 19, 21]. We will discuss additional details on the choice of basis in Section 5.

**Assumption 2 — Availability of a flash photograph.** We resolve the multi-linearity of the unknown parameters by having access to a flash photograph of the scene. In the flash image $I_{\text{f}}$, the intensity observed at pixel $\mathbf{p}$ is given by:

$$I_{\text{f}}^k(\mathbf{p}) = I_{\text{nf}}^k(\mathbf{p}) + \int_\lambda \rho_{\mathbf{p}}(\lambda) S^k(\lambda) \eta_{\text{f}}(\mathbf{p}) \ell_{\text{f}}(\lambda) d\lambda, \quad (4)$$

where $\eta_{\text{f}}(\mathbf{p})$ denotes the shading at $\mathbf{p}$ induced by the flash, and the spectra of the flash $\ell_{\text{f}}$ is assumed to be known via a calibration process. Further, under the reflectance and illumination subspace modeling above, we can write

$$I_{\text{f}}^k(\mathbf{p}) = I_{\text{nf}}^k(\mathbf{p}) + \mathbf{a}_{\mathbf{p}}^\top E^k \eta_{\text{f}}(\mathbf{p}) \mathbf{f}, \quad (5)$$

where $\mathbf{f}$ denotes the illumination coefficients for the flash spectra. We now derive a novel constraint that encodes both the illuminant spectra as well as their shadings at each pixel.

### 3.2. The Hull Constraint

The centerpiece of our approach is a novel reflectance-invariant condition that we call the Hull Constraint. The hull constraint is derived by performing the following three operations (see Figure 2 for a visual guide).

**Step 1 — Estimate the pure flash image.** The pure-flash image $I_{\text{pf}}$ is obtained by subtracting the no-flash image from the flash image:

$$I_{\text{pf}}^k(\mathbf{p}) = I_{\text{f}}^k(\mathbf{p}) - I_{\text{nf}}^k(\mathbf{p}) = \mathbf{a}_{\mathbf{p}}^\top E^k \eta_{\text{f}}(\mathbf{p}) \mathbf{f}. \quad (6)$$

**Step 2 — Solve for reflectance coefficients.** We now have 3 intensity measurements — one per color channel — at $\mathbf{p}$, and 3 unknowns for $\alpha_\mathbf{p} = \eta_f(\mathbf{p})\mathbf{a}_\mathbf{p}$. This enables us to solve for $\alpha_\mathbf{p}$, which corresponds to the reflectance coefficients up to a per-pixel scale $\eta_f(\mathbf{p})$.

**Step 3 — Estimate $\Gamma(\mathbf{p})$.** Since $\frac{\alpha_\mathbf{p}}{\|\alpha_\mathbf{p}\|} = \frac{\mathbf{a}_\mathbf{p}}{\|\mathbf{a}_\mathbf{p}\|}$, we can substitute $\alpha$ to express (3) as:

$$I_{\text{nf}}^k(\mathbf{p}) = \|\mathbf{a}_\mathbf{p}\| \left(\frac{\alpha_\mathbf{p}^T}{\|\alpha_\mathbf{p}\|}\right) E^k \sum_{i=1}^{N} \eta_i(\mathbf{p})\mathbf{b}_i. \quad (7)$$

As before, we are able to solve for $\beta(\mathbf{p})$, defined as

$$\beta(\mathbf{p}) = \|\mathbf{a}_\mathbf{p}\| \sum_{i=1}^{N} \eta_i(\mathbf{p})\mathbf{b}_i. \quad (8)$$

Normalizing $\beta(\mathbf{p})$ gives us $\Gamma(\mathbf{p}) = \beta(\mathbf{p})/\|\beta(\mathbf{p})\|$ that is invariant to the reflectance. We can now state the Hull constraint, which is the main contribution of this paper.

**Proposition 1** (The Hull Constraint). *The term $\Gamma(\mathbf{p})$ lies in the conic hull of the coefficients $\{\mathbf{b}_1, \ldots, \mathbf{b}_N\}$, i.e.,*

$$\boxed{\Gamma(\mathbf{p}) = \frac{\beta(\mathbf{p})}{\|\beta(\mathbf{p})\|} = \sum_{i=1}^{N} z_i(\mathbf{p})\mathbf{b}_i, \quad z_i(\mathbf{p}) \geq 0.} \quad (9)$$

The relative shading term $z_i(\mathbf{p})$ is defined as

$$z_i(\mathbf{p}) = \frac{\eta_i(\mathbf{p})}{\|\sum_j \eta_j(\mathbf{p})\mathbf{b}_j\|}. \quad (10)$$

This term captures the fraction of the shading at a scene pixel that comes from one light source, relative to all the light sources, hence the term *relative* shading. Further, $\Gamma(\mathbf{p})$ belongs to $\mathbb{S}^2$ space since it is unit-norm.

The key insight of the Hull constraint is that $\Gamma(\mathbf{p})$, a quantity that can be estimated from the no-flash/flash image pair, provides an encoding of the illumination coefficients as well as the relative shading. We can hence derive these parameters as well as perform source separation by studying properties of $\Gamma(\mathbf{p})$ over the entire image.

## 4. Source Separation with the Hull Constraint

Recall, from Proposition 1, that $\Gamma(\mathbf{p})$ lies in the conic hull formed by the lighting coefficients $\{\mathbf{b}_1, \ldots, \mathbf{b}_N\}$. We now describe methods to estimate the illuminant spectrum as well as perform source separation from the set $\mathcal{G} = \{\Gamma(\mathbf{p}); \forall \mathbf{p}\}$. Our methods rely on fitting the tightest conic hull to the set $\mathcal{G}$ and identifying the corners of the estimated hull. Along the way, we state sufficient/necessary conditions when the resulting estimates are meaningful. We begin by discussing the identifiability of a light source.

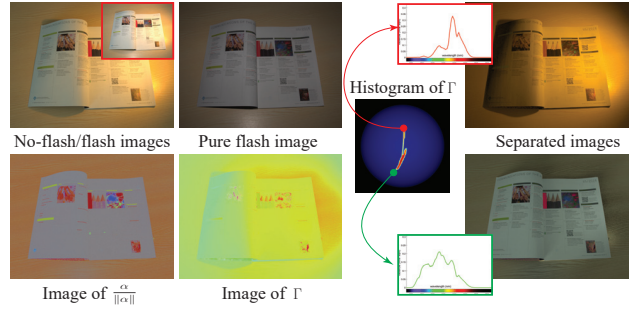

Figure 2. Visualization of our processing pipeline. From the input image pair, we compute the pure flash image as well as values of the $\alpha$ and $\Gamma$ at each pixel. We visualize $\alpha/\|\alpha\|$ and $\Gamma$ as 3-color images by integrating it with the reflectance and illumination bases, respectively, and the camera spectral response. Note that $\alpha/\|\alpha\|$ encodes the scene's reflectance while $\Gamma$, being reflectance-invariant, encodes the shading and illumination. The histogram of $\Gamma$ over the sphere provides the estimate of the illumination spectra as well as the separated images.

### 4.1. Identifiability of a light source

We observe that a light source is identifiable only if its coefficients lie outside the conic hull of the coefficients of the remaining light sources. If this were not that case, then its contribution to a scene point can be explained by the remaining lights. Hence, only light sources whose coefficients lie at corners of the conic hull of $\{\mathbf{b}_1, \ldots, \mathbf{b}_N\}$ are identifiable given the flash/no-flash image pair. Without any loss in generality, we assume that all light sources are identifiable. Therefore, if we can identify the conic hull of the light sources $\mathcal{L} = \text{conic-hull}\{\mathbf{b}_1, \ldots, \mathbf{b}_N\}$, we can estimate the light source coefficients as the corner points of this set. While we do not have an a-priori estimate of $\mathcal{L}$, we can estimate it from the set $\mathcal{G} = \{\Gamma(\mathbf{p}); \forall \mathbf{p}\}$. Recall that, from Proposition 1, $\mathcal{G} \subseteq \mathcal{L}$. We next explore sufficient conditions under which the conic hull of $\mathcal{G}$ is equal to $\mathcal{L}$; when this happens, we can estimate the light source coefficients as the corner points of the conic hull of $\mathcal{G}$.

**Proposition 2** (Presence of "pure" pixels). *Under ideal imaging conditions (absence of noise, non-Lambertian surfaces, etc.), the conic hull of $\mathcal{G}$ is equal to $\mathcal{L}$ if, for each light source, there exists a pixel that is purely illuminated by that light source, or, equivalently,*

$$\forall \mathbf{b} \in \{\mathbf{b}_1, \ldots, \mathbf{b}_N\}, \ \exists \ \Gamma(\mathbf{p}') = \mathbf{b}.$$

When there are pure pixels for each light source, then the set $\mathcal{G}$ will include the illuminant coefficients which are also the corners of the conic hull $\mathcal{L}$. Therefore, the conic hull of $\mathcal{G}$ will be identical to $\mathcal{L}$. Note that pure pixels can be found in shadow regions since shadows indicate the absence of light source(s). The pure pixel assumption is thus satisfied when the scene geometries are sufficiently complex to exhibit a wide array of cast and attached shadows. The more complex the scene geometry, the more likely it is that we satisfy the condition in Proposition 2.

In addition to pure pixels or corners, we can also fit the hull by identifying its edges. Edges of the cone correspond to points that are in the shadow of all but two sources. As with pure pixels, shadows play a pivotal role in recovering the hull from its edges.

### 4.2. Estimating illuminant coefficients

Given the set $\mathcal{G}$, the number of identifiable light sources is simply the number of corners in the tightest conic hull. Hence, we expect the set $\mathcal{G}$ to be concentrated about a point when there is a single light source, an arc with two sources, and so on (see Figure 3). We can use specialized techniques to estimate the parameters in each case (see detailed pseudo-code in the supplemental material).

- $N = 1$ — While not particularly interesting in the context of source separation, we use the robust mean of $\mathcal{G}$ to be the coefficients of the single light source.

- $N = 2$ — We use RANSAC to robustly estimate the arc on $\mathbb{S}^2$ with maximum inliers. The end points of this arc are associated with the illuminant coefficients; this estimate will correspond to the true coefficients if there were "pure pixels" in the no-flash photograph for each of the light sources.

- $N = 3$ — We project the set $\mathcal{G}$ onto the tangent plane at its centroid and fit the triangle with least area onto the projected points. Fitting polyhedra onto planar points has been extensively studied in computational geometry [34, 18, 3, 32, 35]. We use the method in Parvu et al. [35] to determine the triangle and the associated vertices.

- $N \geq 4$ — The procedure used for three light sources can potentially be applied to higher number of sources. However, as we will see next, even if we can estimate the lighting coefficients, source separation with a three-color camera cannot be performed when $N \geq 4$.

For the results in the paper, we manually specify the number of light sources (typically, 2 or 3) and use the corresponding algorithm to extract the corners. Given the estimated lighting coefficients $\{\widehat{\mathbf{b}}_1, \ldots, \widehat{\mathbf{b}}_N\}$, we can estimate the relative shading at each pixel.

### 4.3. Estimating the relative shading

Given $\Gamma(\mathbf{p})$ and estimates of the lighting coefficients $\{\widehat{\mathbf{b}}_1, \ldots, \widehat{\mathbf{b}}_N\}$, we simply solve the linear equations in (9) under non-negativity constraints to estimate the relative shading $\{z_i(\mathbf{p}), i = 1, \ldots, N\}$. It is easily shown that there is a unique solution when $\Gamma(\mathbf{p}) \in \text{conic-hull}\{\widehat{\mathbf{b}}_1, \ldots, \widehat{\mathbf{b}}_N\}$ and $N \leq 3$ (see supplemental material). When $N > 3$, we can obtain multiple solutions to the relative shading — a limitation that stems from using 3-color cameras.

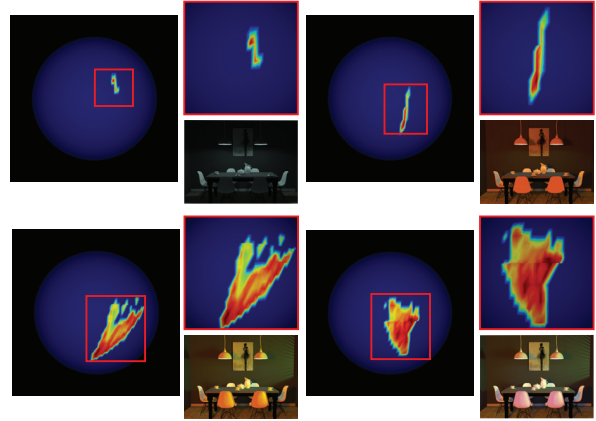

Figure 3. Visualization of $\mathcal{G}$ as a histogram on $\mathbb{S}^2$ for scenes with different number of sources. As expected, the histogram takes progressively complex shapes with increasing number of sources.

### 4.4. Lighting separation

Once we have the illumination coefficients $\{\widehat{\mathbf{b}}_1, \ldots, \widehat{\mathbf{b}}_N\}$ and the relative shading $\{\widehat{z}_i(\mathbf{p})\}$, we can separate the no-flash photograph into $N$ photographs. Specifically, for the $j$-th light source, we would like to estimate
$$I_{\text{sep},j}^k = \mathbf{a}_{\mathbf{p}}^\top E^k \eta_j(\mathbf{p}) \mathbf{b}_j.$$
An estimate of this image is obtained as
$$\widehat{I}_{\text{sep},j}^k(\mathbf{p}) = \|\beta\| \alpha_{\mathbf{p}}^\top E^k \widehat{z}_j(\mathbf{p}) \widehat{\mathbf{b}}_j. \tag{11}$$

## 5. Evaluation and Applications

We characterize the performance of the proposed methods by evaluating light separation and showcasing its potential in a number of applications.

**Capture setup for real data.** The flash/no-flash images were captured using a Nikon D800 and a Speedlight SB-800 flash, with the camera mounted on a tripod and operated under aperture-priority mode. The images were captured in raw format and demosaiced under a linear response using DCRaw [1]. Finally, the flash spectrum was assumed to be flat, i.e., $\ell_f(\lambda)$ in (4) was assumed to be a constant.

**Selection of reflectance and illumination bases.** We used the measured database for reflectance [25] and illumination [2] to learn two three-dimensional subspaces, one each for reflectance and illumination. All the results in this paper were obtained with the same pair of bases, which we learned using a weighted PCA model, with the camera spectral response providing the weights. We observed that this

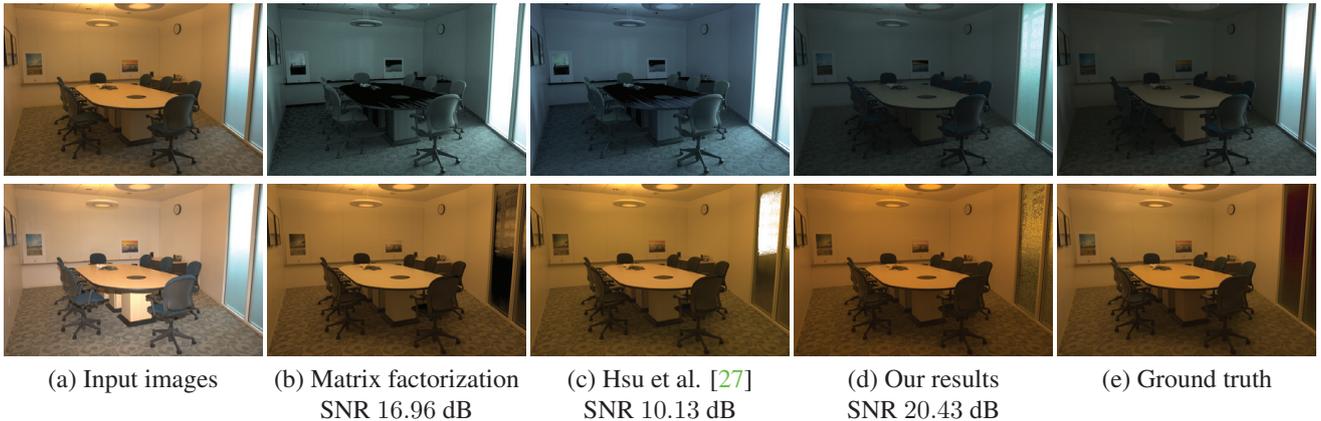

(a) Input images    (b) Matrix factorization    (c) Hsu et al. [27]    (d) Our results    (e) Ground truth
SNR 16.96 dB      SNR 10.13 dB      SNR 20.43 dB

Figure 4. We separate a no-flash image (a) into two components and compare with matrix-factorization (b) and and Hsu et al. [27] (c). Compared to the ground truth images, we can see that matrix factorization produces noisy colors (see the painting on the left), while Hsu et al. [27] produce an incorrect estimate of light color and shading. Our result (d) closely mimics the actual captured results.

technique outperformed an unweighted PCA as well as the joint learning of subspaces [12]. The supplemental material provides a detailed evaluation on synthetic scene with comparisons to alternate strategies.

**Pruning $\mathcal{G}$.** To reduce effects of measurement noise and model mismatch, we build a histogram of $\mathcal{G}$ by dividing the sphere into $100 \times 100$ bins and counting the occurrence of $\Gamma(\mathbf{p})$ in each bin. We remove points in sparsely populated regions; typically, points in bins that have less than 100 pixels are removed from $\mathcal{G}$.

### 5.1. Evaluation of lighting separation

We report the performance of our source separation technique on a wide-range of real-world scenes. The accompanying supplementary material contains additional results and comparisons.

**Synthetic experiments.** The supplemental material also provides rigorous evaluation of the source separation technique on realistically-rendered scenes using the MITSUBA rendering engine [29]. As a summary, for the two light sources scenario, the normalized mean square error in separated images is less than $10^{-3}$ when the light sources coefficients are more than $10°$ apart.

**Scenes with two lights.** In Figure 4, we demonstrate our technique on the scene with two lights sources and compare with ground truth captures. Ground truth photographs were obtained by turning off the indoor light sources to obtain the outdoor illuminated scene and then subtracting this from the no-flash image to obtain the photograph with respect to the indoor illumination. We also compare against a simple non-negative matrix factorization (NNMF) as well as the technique proposed in Hsu et al. [27]. Naively applying NNMF to the no-flash image leads to the loss of the colors. Hsu et al. [27] use the no-flash photograph to estimate the relative contribution of the light sources by introducing

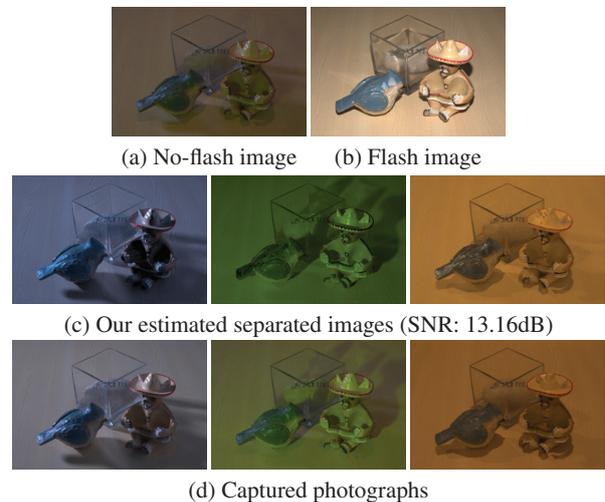

(a) No-flash image    (b) Flash image

(c) Our estimated separated images (SNR: 13.16dB)

(d) Captured photographs

Figure 5. We evaluate our technique on scenes with mixtures of three lights and compare with the ground truth image. Our technique is able to capture both the color and the shading for each of these sources and produce results similar to the ground truth.

restrictive assumptions on the scene as well as the colors of the illuminants; while we manually selected the light colors to guide the reconstruction of this technique, there are numerous visual artifacts due to the use of strong scene priors. In contrast, our technique produces results that closely resemble to the actual captured photographs, indicating its robustness and effectiveness.

**Scenes with three lights.** The proposed technique is, to our knowledge, the first to demonstrate three light source separation. In Figure 5, we compare our technique to the ground truth on scenes with three lights. The scene is illuminated under warm indoor lighting, a green fluorescent lamp and cool skylight. Our lighting separation scheme produces visually pleasing results with shadows and shadings that are consistent with those observed in the ground truth.

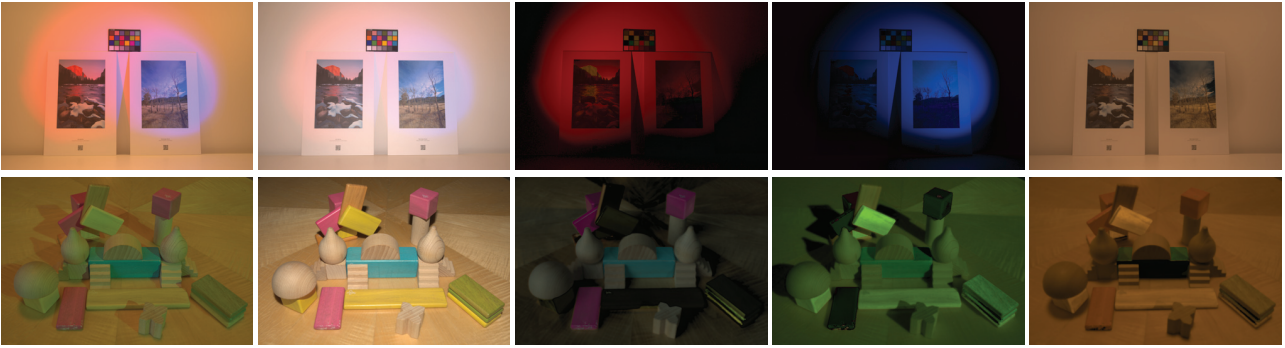

(a) No-flash images    (b) Flash images    (c) Estimated separated images

Figure 6. We evaluate our technique on scenes with three lights. (top row) We capture an image under warm indoor LED lights and two LED lights with red and blue filter, respectively. Our technique is able to estimate separated results that capture this complex light transport. (bottom row) We image a scene under warm indoor lighting, a green fluorescent lamp and cool skylight. Our separation results capture both the color and the shading for each of these sources.

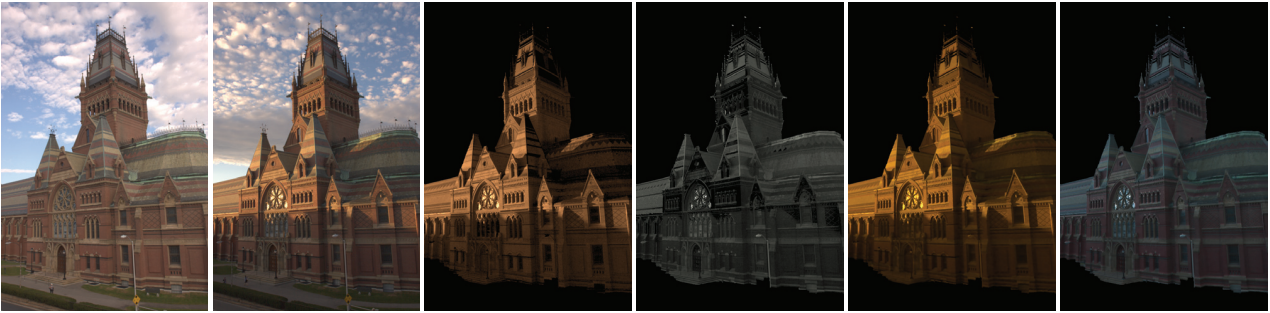

Frame 40 (pure flash)    Frame 16    Method of Prinet et al. [38]    Our results

Figure 7. Sun- and Sky-light separation. We use photo on a cloudy day as the pure flash image. Note that the Sun being a directional light source casts sharp shadows onto the scene, while the Sky being an area light does not induce shadows. As can be seen from the separated images, our algorithm is able to produce good results with convincing color and shading attributes for both sources in the scene.

Figure 6 showcases separation on two additional scenes. For the scene in the top row, our technique for estimating lighting coefficients fails due to lack of shadows; to obtain the separation, we had to manually pick the corners of $\mathcal{G}$ to estimate the illumination coefficients.

### 5.2. Applications

Source separation of the form proposed is invaluable in many applications. We consider five distinct applications: white balancing under mixed illumination, manipulation of camera spectral response, post-capture editing of illuminant spectrum and brightness, sun/sky-light separation and two-shot photometric stereo. Due to space constraints, we cover white balancing and camera response manipulation in the supplemental material.

**Sunlight and skylight separation.** An interesting application of two-light source separation is in outdoor time lapse videos where it is often necessary to separate direct sunlight from indirect skylight. Figure 7 showcases the performance of light separation technique on an outdoor scene.

We identify a photograph with cloudy sky, where there is no direct sunlight and the entire scene is lit only by the skylight, as a pure flash photograph. Since our technique does not make any assumptions about the nature of the flash illumination, we use skylight in place of the flash light. Also note that skylight changes its color and intensity significantly during the course of the day. Given this pure flash photograph, our separation scheme is able to produce the results closely resemble to the manner of the sky and the sun illumination. We compare our method with the video-based work of Prinet et al. [38] on the time-lapse video sequence. While the method by Prinet et al. does not require the pure flash image, it assumes that the colors of the illuminants will not change which leads to artifacts in the separated images.

**Post-capture manipulation of light color and brightness.** Given the separated results, we can adjust the brightness as well as the spectrum of a particular light. Specifically, we can produce the photograph

$$\widetilde{I} = \sum_j \|\beta\|_2 \alpha_\mathbf{p}^T E^k \widehat{z}_j(\mathbf{p}) \mu_j \widetilde{\mathbf{b}}_j, \quad (12)$$

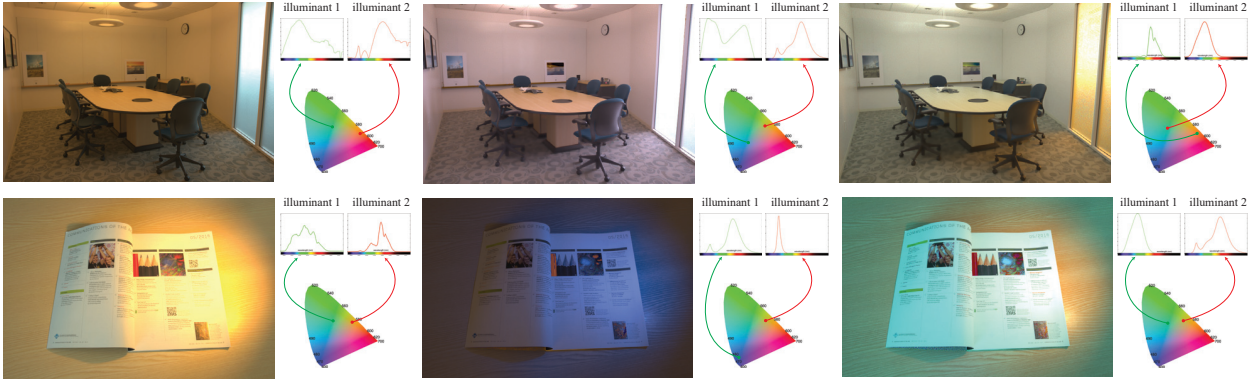

(a) No-flash image            (b) Our light editing results

Figure 8. We separate no-flash images (a) into individual light components, and recolor them to create photo-realistic results with novel lighting conditions (b). We show the novel spectral distribution as well as the CIE plots for the light sources. Note how our method changes the color and brightness of each light while realistically retaining all shading effects.

where $\widetilde{\mathbf{b}}_j$ denotes the adjusted illumination coefficients and $\mu_j$ denotes the changes in the brightness. Figure 8 shows two examples of editing the light color and brightness for the captured no-flash images. We experiment by adjusting the parameters $\mu_j$ and $\widetilde{\mathbf{b}}_j$ in (12). The rendered photographs are both visually pleasing and photorealistic in their preservation of shading and shadows.

**Flash/no-flash photometric stereo.** Photometric stereo [44, 40] methods aim to surface shape (usually normals) of an object from images obtained from a static camera under varying lighting. For Lambertian objects, this requires a minimum of three images. Recently, techniques have been proposed to do this from a single shot where the object is lit by three monochromatic red, green, and blue, directional light sources [9, 11]. However this estimation is still ill-posed and requires additional priors. We propose augmenting this setup by capturing an additional image lit by a flash collocated with the camera. We use our proposed technique for source separation to create three images (plus the pure flash image), at which point we can use standard calibrated Lambertian Photometric Stereo to estimate surface normals. As shown in Figure 9 this leads to results that are orders of magnitude more accurate than the state-of-the-art technique [11]. More comparisons can be seen in the supplementary material.

## 6. Conclusions

In this paper, we have shown that capturing an additional image of a scene under flash illumination, allows us to separate the no-flash image into image corresponding to illuminants with unique spectral distribution. Our technique is based on a novel colorimetric analysis that derives the conditions under which this is possible. We have also shown that this ability to analyze and isolate lights in turn leads to state-of-the-art results on tasks such as white balancing,

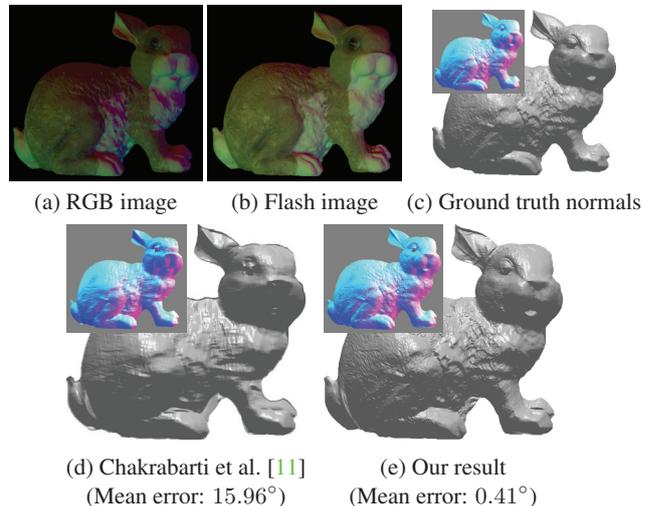

(a) RGB image    (b) Flash image    (c) Ground truth normals

(d) Chakrabarti et al. [11]    (e) Our result
(Mean error: $15.96°$)    (Mean error: $0.41°$)

Figure 9. Results on two-shot captured photometric stereo of real objects. We show estimated normal map for our technique as well as that of single-shot method of Chakrabarti et al. [11]. We also include the mean of the angular errors for the estimated surface normals. Our technique produces surface normals with very low angular errors.

illumination editing, and color photometric stereo. We believe that this is a significant step towards true post-capture lighting control over images.

## 7. Acknowledgment

The authors thank Prof. Shree Nayar for valuable insights on an earlier formulation of the ideas in the paper. Hui and Sankaranarayanan acknowledge support via the NSF CAREER grant CCF-1652569, the NGIA grant HM0476-17-1-2000, and a gift from Adobe Research.